\documentclass[final]{cvpr}
\usepackage{makecell}
\usepackage{times}
\usepackage{epsfig}
\usepackage{subfig} 
\usepackage{graphicx}
\usepackage{amsmath}
\usepackage{amssymb}

\usepackage{amsmath,amsfonts,bm}

\def\eqref#1{equation~\ref{#1}}

\def\1{\bm{1}}

\def\ve{{\bm{e}}}

\def\vp{{\bm{p}}}

\def\vs{{\bm{s}}}

\def\vx{{\bm{x}}}
\def\vy{{\bm{y}}}

\DeclareMathAlphabet{\mathsfit}{\encodingdefault}{\sfdefault}{m}{sl}
\SetMathAlphabet{\mathsfit}{bold}{\encodingdefault}{\sfdefault}{bx}{n}

\def\vggloss{\mathcal{L}_{\text{VGG}}}
\def\vggfaceloss{\mathcal{L}_{\text{VGGFace}}}

\def\render{\vy}
\def\shape{\vs}
\def\flame{F}
\def\pose{\vp}
\def\expression{\ve}
\def\lighting{\iota}

\def\style{\zeta}

\def\tgt{\text{tgt}}

\def\renderer{R}

\usepackage[pagebackref=true,breaklinks=true,colorlinks,bookmarks=false]{hyperref}

\begin{document}

\title{MorphGAN: One-Shot Face Synthesis GAN for Detecting Recognition Bias}

\author{Nataniel Ruiz$^1$
\quad
Barry-John Theobald$^2$
\quad
Anurag Ranjan$^2$ \\
Ahmed Hussein Abdelaziz$^2$
\quad
Nicholas Apostoloff$^2$ \\ \\

$^1$Boston University, Boston, MA 
\quad $^2$Apple, Cupertino, CA\\

{\tt\small nruiz9@bu.edu \quad \{barryjohn\_theobald, anuragr, hussenabdelaziz, napostoloff\}@apple.com}

}
\maketitle

\begin{abstract}
To detect bias in face recognition networks, it can be useful to probe a network under test using samples in which only specific attributes vary in some controlled way.  However, capturing a sufficiently large dataset with specific control over the attributes of interest is difficult. In this work, we describe a simulator that applies specific head pose and facial expression adjustments to images of previously unseen people.  The simulator first fits a 3D morphable model to a provided image, applies the desired head pose and facial expression controls, then renders the model into an image.  Next, a conditional Generative Adversarial Network (GAN) conditioned on the original image and the rendered morphable model is used to produce the image of the original person with the new facial expression and head pose. We call this conditional GAN -- {MorphGAN}.

Images generated using MorphGAN conserve the identity of the person in the original image, and the provided control over head pose and facial expression allows test sets to be created to identify robustness issues of a facial recognition deep network with respect to pose and expression.  Images generated by MorphGAN can also serve as data augmentation when training data are scarce. We show that by augmenting small datasets of faces with new poses and expressions improves the recognition performance by up to 9\% depending on the augmentation and data scarcity.
\end{abstract}

\section{Introduction}
\label{sec:introduction}

\begin{figure*}[t]
    \centering
    \includegraphics[width=0.75\textwidth]{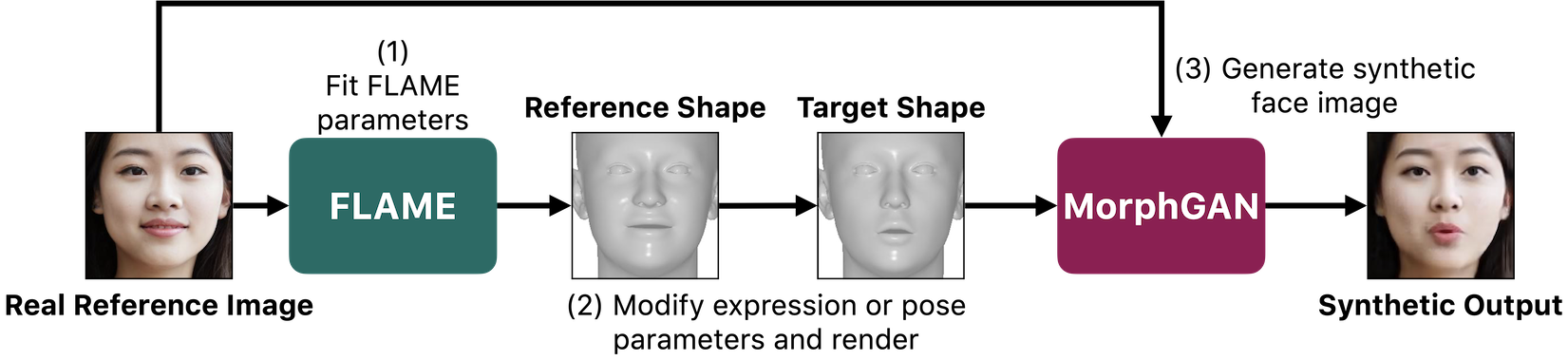}
\caption{Our one-shot face synthesis pipeline with controllable expression and pose. (1) We estimate the shape, pose and expression parameters of the FLAME face model.  (2) We modify the parameters that control pose and facial expression, and render the new model instance. (3) Finally, we use the rendered model and original image to generate a new image of the source identity with the desired pose and facial expression.}
\label{fig:pipeline}
\end{figure*}

Face recognition is commonly used to secure personal devices. Training robust face recognition systems \cite{abate20072d} requires sufficiently diverse training data to avoid bias \cite{Garcia_2019}. However, it is difficult to curate a dataset that is large and diverse, and has the explicit consent of all of the people included in the dataset. A solution to the problem of dataset size is to exploit recent advances in generative adversarial networks (GANs) \cite{choe2017face, mokhayeri2020cross}, where realistic facial images are generated using appropriate control parameters \cite{karras2019style, viazovetskyi2020stylegan2}. However, the ability to control GANs is limited. StyleGAN~\cite{karras2019style, viazovetskyi2020stylegan2} can generate a diverse collection of users, but does not provide explicit controls to manipulate facial expressions. HoloGAN~\cite{nguyen2019hologan} provides controls only for head rotations. Zakharov et al.~\cite{zakharov2019few} condition a GAN on reference frames of a given person to manipulate the face using facial landmarks.  The best result are achieved using several reference images. However, there is no direct control for expression or pose.

3D morphable models (3DMM) \cite{history3dmm} are an alternative to GANs.  3DMMs use complex controls, typically derived using principal components analysis, to produce high-quality animation. However, images generated using 3DMMs lack realism. Ghosh et al.~\cite{ghosh2020gif} condition a GAN on FLAME~\cite{flame} to provide explicit controls for head-rotations and facial expressions. However, these controls can manipulate only the images generated by the GAN, and cannot manipulate an image of a real face.

We present an approach that provides controlled manipulation of head pose and facial expression of any given face in an image whilst preserving the identity. Thus, our method can be used to increase the diversity of faces in a dataset to improve the robustness of face recognition models.  This is especially important in situations where the number of samples for an individual is limited. 

Given an image containing a face of a person of interest, we estimate the FLAME parameters to obtain the 3D geometry of the face. These parameters can then be used to manipulate the 3D geometry over different head poses and facial expressions. Our conditional GAN takes as input an image and the target 3D geometry, and generates an output image of the identity in the target head pose and with target facial expression.  This allows us to perform sensitivity tests of a facial recognition model by generating data samples of specific identities whilst varying only a single attribute \cite{webster2018visual}.  This level of control over the synthesized face allows us to detect and evaluate bias in the facial recognition model.

The data generated by our face synthesis network can also be used to augment datasets to include new poses and facial expressions for existing identities. We show that the augmentations generated by our model improve the result for a facial recognition system by up to 9\% when limited data are available.

{\bf Contributions.}
In summary, our main contributions are:
\begin{enumerate}
\item{\emph{MorphGAN}: a conditional GAN that generates face images with a desired identity, pose and expression by conditioning on a 3D face model and a single reference image of the desired identity.}
\item{A sensitivity test to detect (pose/expression) bias of a face recognition network using \emph{MorphGAN} generated images of individuals used to train the  network.}
\item{Improved recognition performance using \emph{MorphGAN} images to augment a small training dataset by expanding the variation of pose and expression.}
\end{enumerate}

\section{Related Work}
\label{sec:related_work}

Early work on generating realistic face images used 3DMMs~\cite{blanz1999morphable}.  3DMMs provide a dense representation of face shape, and a corresponding 2D texture to represent appearance.  Furthermore, the model parameters required to represent a specific face can be estimated from an image automatically, and then later used to manipulate the shape and appearance of the rendered face. Kim et al.~\cite{kim2018DeepVideo} used a deep convolutional neural network to refine a rendered 3DMM approximation of a desired image. Geng et al.~\cite{geng20193d} train networks conditioned on expression coefficients to generate the shape and the texture separately. Ranjan et al.~\cite{coma} use a convolutional mesh autoencoder to learn a representation of 3D shapes under extreme expressions.

Alternatively, deep generative methods can be applied to the problem of facial image synthesis.  For example, X2Face~\cite{wiles2018x2face} modifies an input image of a face according to a \emph{driving} source, which could be a different facial image or audio.  The approach works well for small changes in head pose or facial expression, but suffers artifacts due to warping if the required transform is large.  GANs, and the many variants~\cite{goodfellow2014generative, isola2017image, karras2018progressive, wang2018high, wang2018video}, have received increasing attention because of the high quality images that they can produce. One approach is to condition the network on multiple images of the target person, and input the 2D landmarks of the face in the desired pose and facial expression~\cite{zakharov2019few}.  The network then learns to generate the desired image from the inputs.  However, limitations are a lack of explicit control for pose and expression --- the specific landmarks must be provided, and \emph{identity leak} can occur if the landmarks are from a different identity. GANimation~\cite{pumarola2018ganimation} uses a generator conditioned on action units (AUs) of the Facial Action Coding System (FACS) to generate an attention map to control which areas of the source face need to be modified to transform to a target expression.  The advantage of using AUs over landmarks is that AUs provide more intuitive controls for the facial expression.  However, a limitation of GANimation is the head pose cannot be altered.

The U.S. Department of Commerce released a report showing that contemporary commercial face recognition algorithms exhibit false positive rates that are highest in West and East African and East Asian people, and are lowest in Eastern European individuals~\cite{grother2019face}. Moreover, gender~\cite{buolamwini2018gender} and other sources of bias  \cite{nagpal2019deep, cavazos2020accuracy} have also been detected in face recognition software.  One approach to mitigate the effects of bias is to use data augmentation to create samples to re-balance the training data.  However, care is required to ensure that mitigating one source of bias does not introduce bias with respect to other attributes \cite{Alvi_2018_ECCV_Workshops}, including non-obvious sources, e.g.\ image quality \cite{Cavazos_2020}.  To identify sources of bias, Kortylewski et al.\ \cite{kortylewski2018empirically, kortylewski2019analyzing} used a 3DMM to generate data to understand changes in facial recognition rate as a function of a specific facial attribute.  However, the approach is based on 3DMMs, which lack the realism of GAN generated images.  An interesting finding in \cite{Robinson_2020_CVPR_Workshops} using the \emph{balanced faces in the wild} dataset is that not all data should be considered equal in terms of fairness.  For example, the same decision threshold is typically used for all data, which usually hurts certain subgroup(s) even if the overall \emph{best} result is obtained.  One particular measure of fairness is parity \cite{bellamy2019ai} where performance is equal across all subgroups.  This is a desirable property and can be achieved using a decision threshold that varies by subgroup.

In this paper we use present a novel GAN-based face image generator with interpretable controls that allow precise manipulation of facial attributes.  We can use this generator to probe a facial recognition network to understand sources of bias as well as create additional samples to re-balance the training data.

\section{Methods}\label{sec:method}

\begin{figure*}[t]
    \subfloat[]{\includegraphics[scale=0.145]{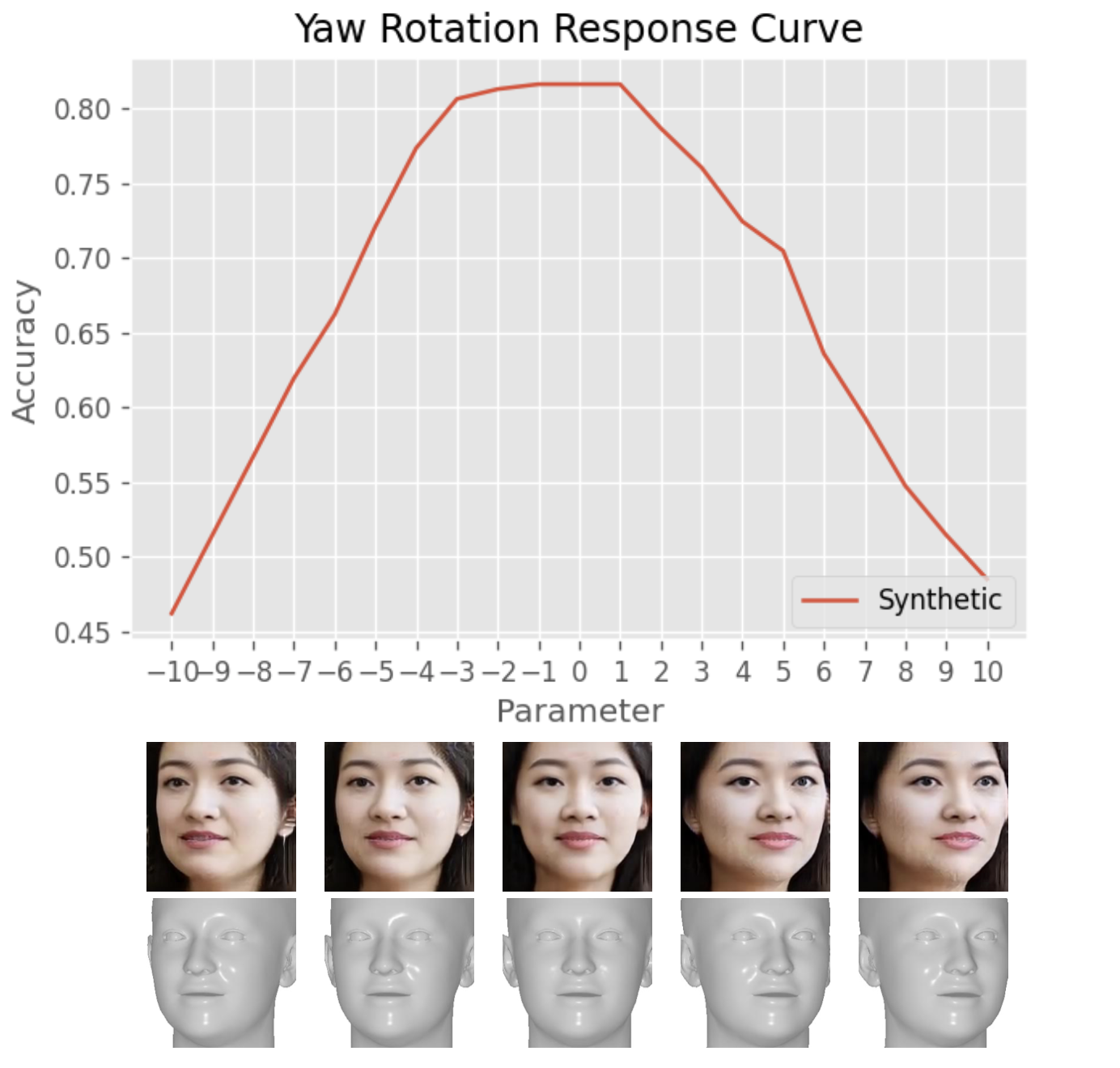}}
    \subfloat[]{\includegraphics[scale=0.145]{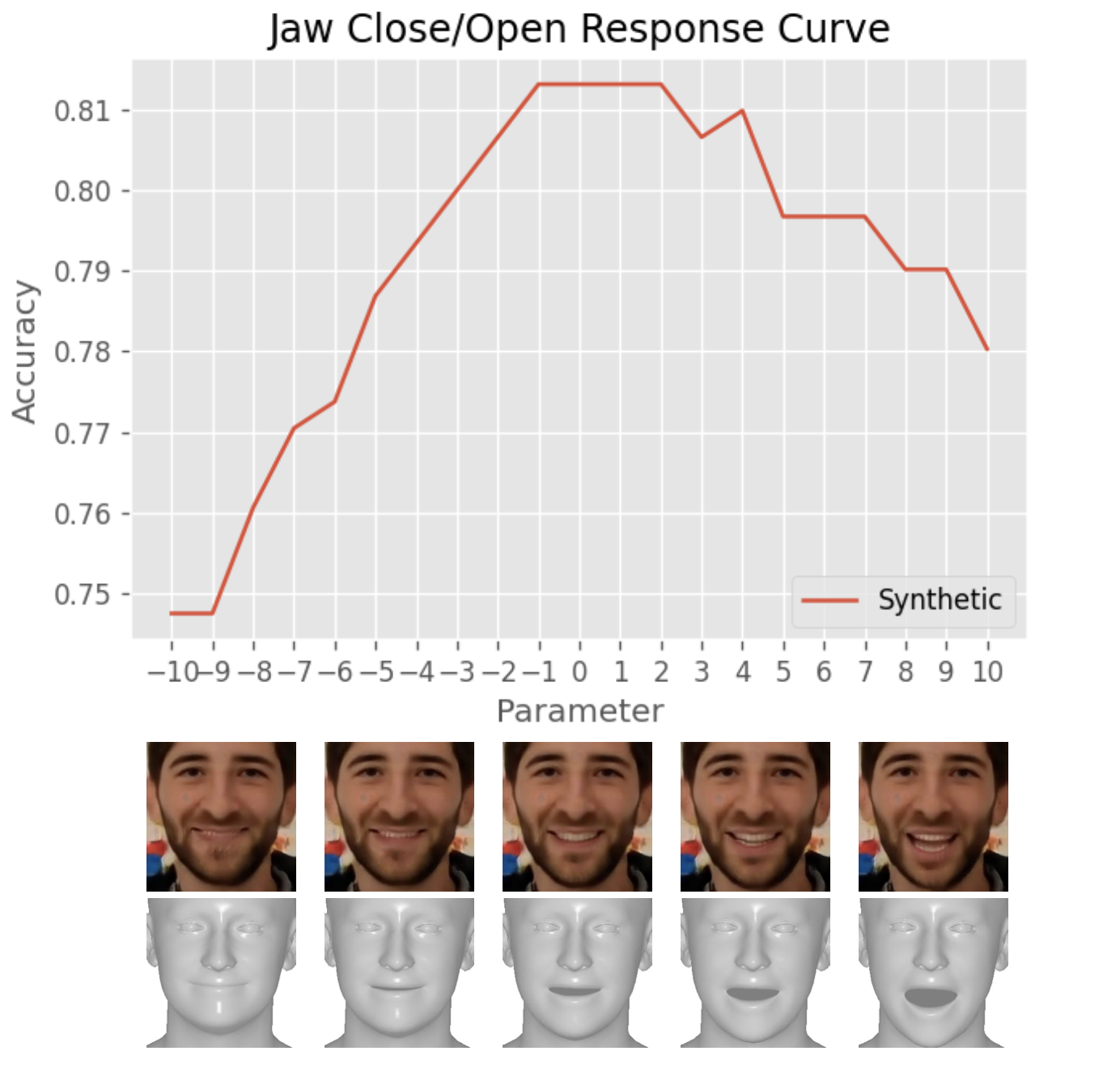}}
    \subfloat[]{\includegraphics[scale=0.145]{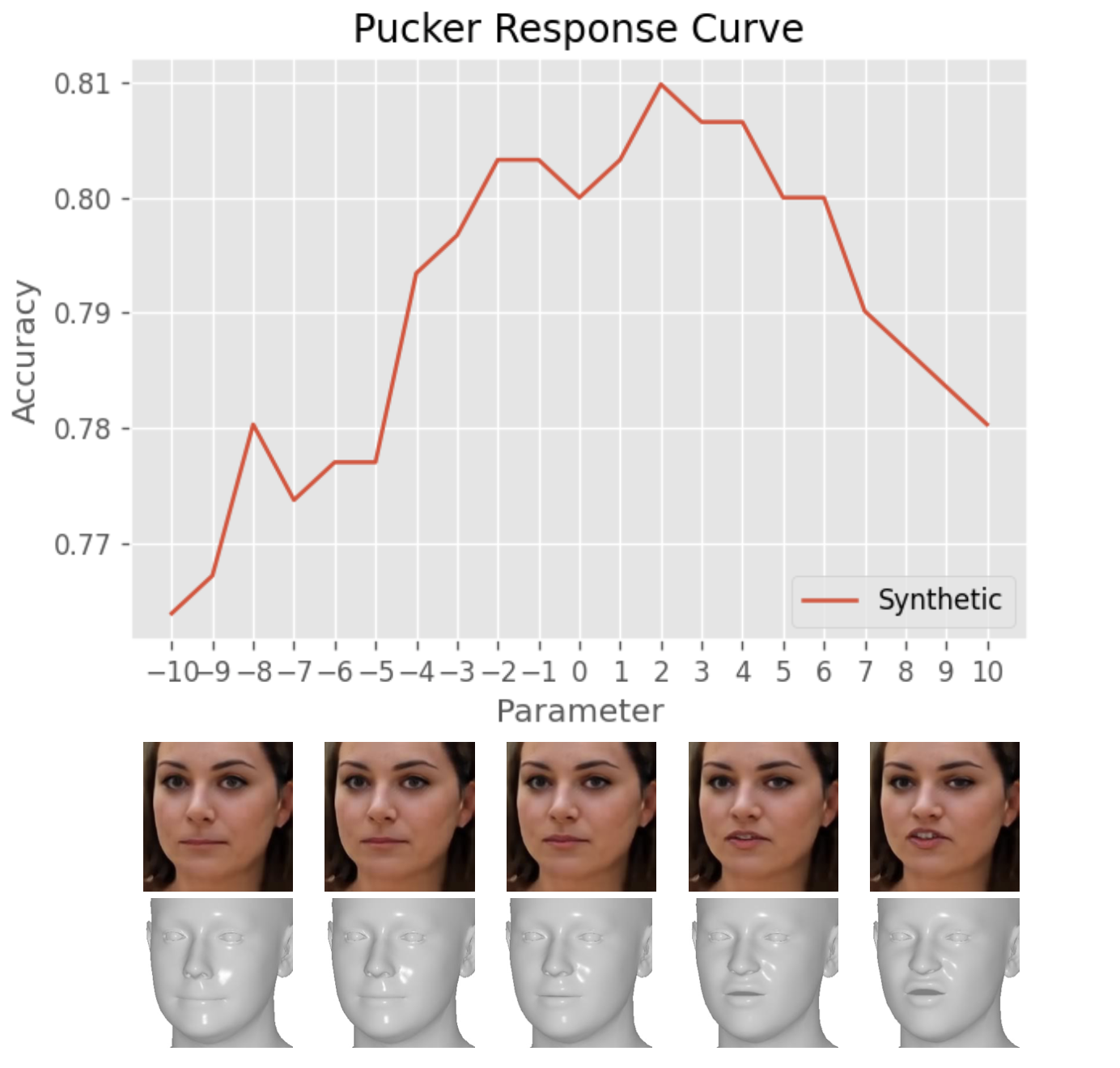}}
    \subfloat[]{\includegraphics[scale=0.145]{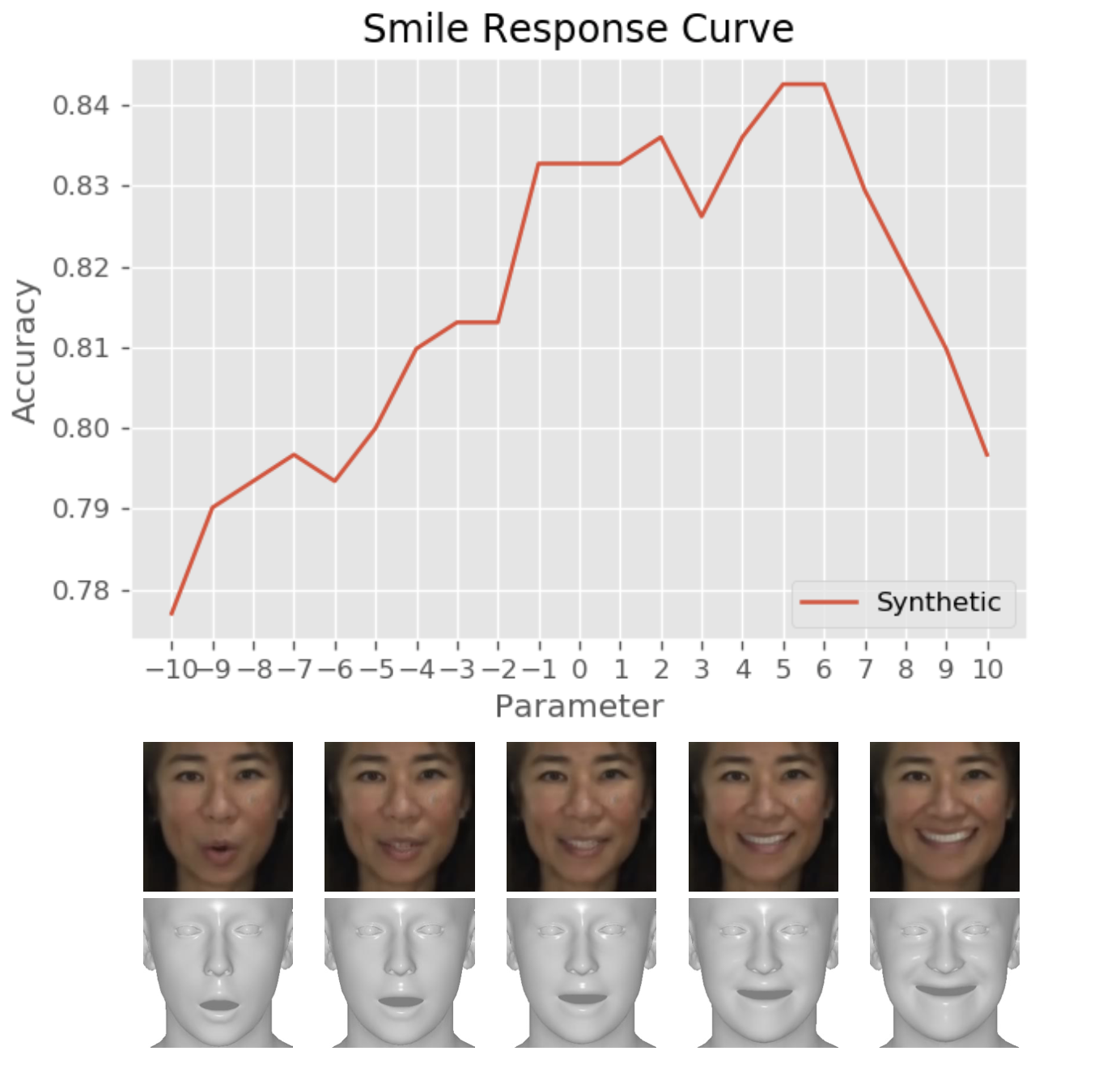}}
    \subfloat[]{\includegraphics[scale=0.145]{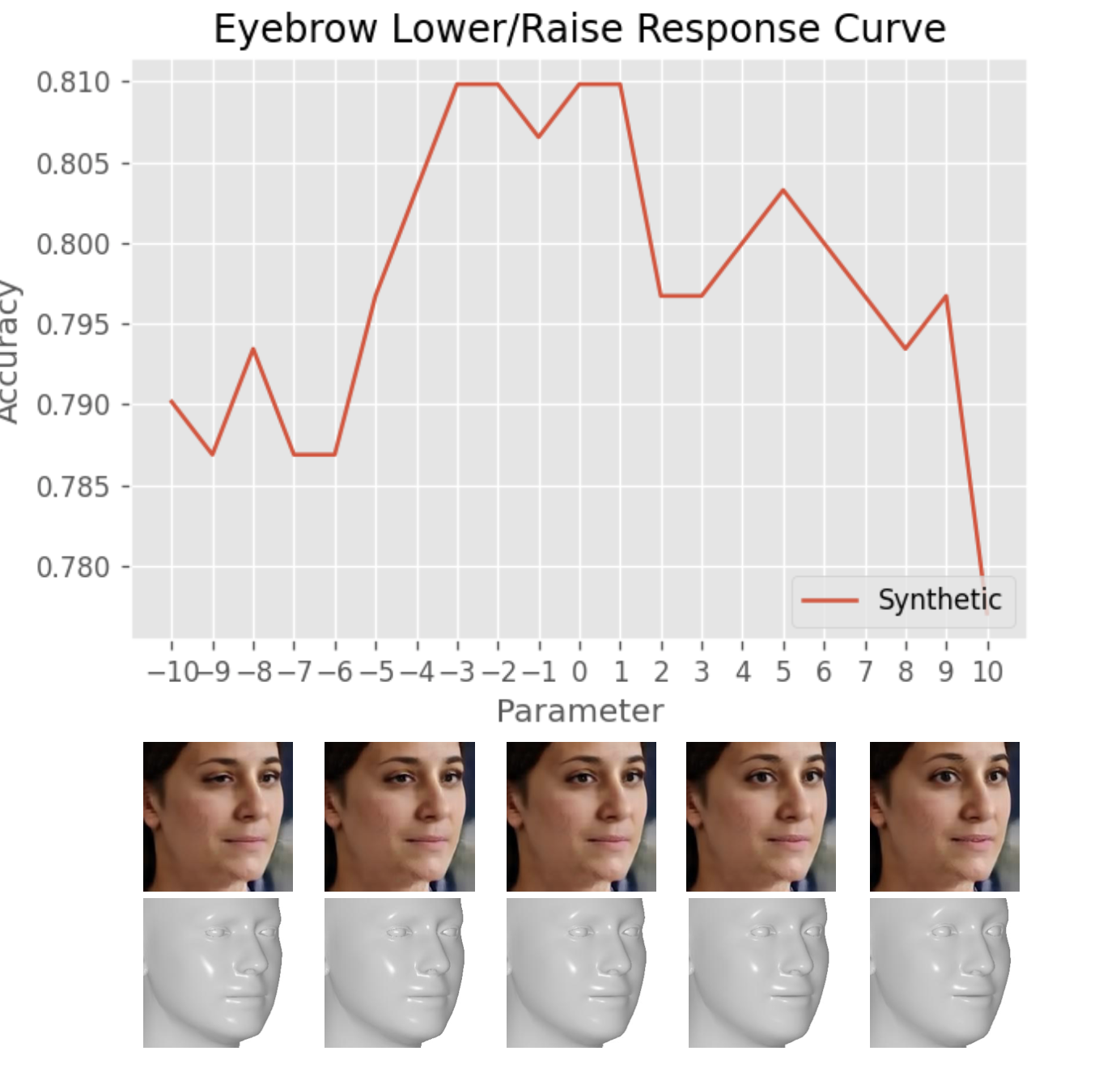}}
\caption{Diagnosis curves for different expression and pose changes on a subset of the AVSpeech Test dataset. The examples shown under the curves illustrate the specific pose or expression change, and the images are generated using MorphGAN.  Samples (a), (c) and (e) are identities generated by StyleGAN2 trained on the FFHQ dataset, and samples (b) and (d) are subjects that provided their images to the authors with express permission to modify the facial expressions.}
\label{fig:diagnosis_experiment}
\end{figure*}

We now describe our generative face model, MorphGAN, that synthesizes faces with varying head pose and facial expression conditioned on a single image of any identity. We will use this model to test the accuracy of a face recognition model over variations in head pose and facial expression.

\subsection{Face Synthesis}
Our pipeline for face synthesis, shown in Figure~\ref{fig:pipeline}, first generates a representation of the target face shape, and then uses this shape representation along with a reference image to render a realistic face image with the desired expression and pose.  This separation of shape and appearance generation has been demonstrated to be effective \cite{chen:2019:hierarchical,chen:2019:photo-realistic,zhenglin:2019:face}. Our approach is to use FLAME~\cite{flame} to decompose a face shape into the inherent (identity) shape, pose and expression parameters.  During training, we use image pairs, a \emph{reference} input and a \emph{target} output.  For each image, we extract the 2D facial landmarks and then fit the FLAME model using these landmarks to estimate the FLAME parameters. Our network then learns to generate the \emph{target} image given the \emph{reference} input and \emph{target} FLAME parameters.  An overview of the pipeline is shown in Figure~\ref{fig:pipeline}.  At test-time, we fit the FLAME parameters to a test image, and then modify the pose and expression parameters, whilst keeping the shape parameters constant.  This updated head model is then used to render the output image of the same person under new head pose and facial expression.

\paragraph{3D Morphable Face Model}

To control the facial expression and head pose in a photorealistic synthesized image, the generator must be conditioned on both the identity of the person and the desired facial expression and head pose. The representation of expression and pose are important for training the network, and we use a rendered image of a FLAME model~\cite{flame}, as shown in Figure~\ref{fig:pipeline}. In particular, we generate rendered images, $\render$, of the FLAME model, $\flame$, that is instantiated using the desired shape, $\shape$, expression, $\expression$, and pose, $\pose$, parameters, and rendered under a certain lighting, $\lighting$, using a renderer, $\renderer$: 
\begin{equation}
\vy = \renderer(\flame(\shape, \expression, \pose), \lighting).
\label{eqn:shape_model}
\end{equation}
Given an image, we detect the facial landmarks and fit the 3D model to these landmarks to recover $\shape$, $\expression$ and $\pose$.  We condition our generator on the original image, and hold $\lighting$ and $\shape$ constant, so that new images of the same person can be generated by varying $\expression$ and $\pose$.

\paragraph{MorphGAN: Rendering Faces}

MorphGAN is a conditional image translation GAN, which allows previously unseen faces to be rendered in new poses and with new facial expressions.  Importantly, MorphGAN requires no fine-tuning or retraining.  The generator is conditioned on a reference face image $\vx_{\text{ref}}$, the rendered target face $\vy_{\text{tgt}}$ and a style vector $\style_{\text{ref}}$ as shown in Figure~\ref{fig:pipeline}.

The MorphGAN generator can be written $G(\vx_{\text{ref}}, \vy_{\text{tgt}}, \style_{\text{ref}})$. The style vector $\style$ is generated by a style encoder network $\style = S(\vx)$.
We use two discriminator networks, a global discriminator $D_{1}(\vx, \vy_{\text{tgt}}, \style_{\text{ref}})$ and a patch discriminator~\cite{isola2017image} $D_{2}(\vx)$, where $\vx$ is either an image generated by $G$ or a real face image from the training dataset, $\vy_{\text{tgt}}$ is the rendered face shape, and $\style_{\text{ref}}$ is the style vector extracted from the reference image. We describe the network architectures in the supplementary material.

The networks are trained with losses as described below.  We use a supervised perceptual loss $\vggloss$, using the activations from layers of an ImageNet pre-trained truncated VGG-19 network~\cite{simonyan2014very}. We apply L1 loss to the activations of layers $\{4, 9\}$ for both the  synthesized image and the target image, given by: 
\small
\begin{multline}
\vggloss(\vx_\text{ref}, \vx_\text{tgt}) = \\ \sum_{i \in \{4, 9\}} M_i ||H_{1}^{(i)}(\vx_{\text{tgt}}) - H_{1}^{(i)}(G(\vx_{\text{ref}}, \vy_{\text{tgt}}, \style_{\text{ref}}))||_1,
\end{multline}
\normalsize
where $M_{4}=\frac{1}{2}$, $M_{9}=1$, and $H_{1}^{(i)}$ are the feature outputs for the i-th layer of the VGG-19 network.  This perceptual loss helps to capture local detail~\cite{gatys2015artistic,gatys2015texture}. To capture higher-level facial features, we use a loss featuring a VGGFace2 pre-trained VGG-13 network~\cite{cao2018vggface2}, where we compute a weighted L1 loss from the activations of convolutional layers $\{10, 13\}$ with respective weights for both the synthesized image and the target image.  We can write this loss as:
\small
\begin{multline}
\vggfaceloss(\vx_\text{ref}, \vx_\text{tgt}) = \\ \sum_{i \in \{10, 13\}} W_i ||H_{2}^{(i)}(\vx_{\text{tgt}}) - H_{2}^{(i)}(G(\vx_{\text{ref}}, \vy_{\text{tgt}}, \style_{\text{ref}}))||_1,
\end{multline}
\normalsize
where $W_{10}=\frac{1}{2}$ and  $W_{13}=1$, and $H_{2}^{(i)}$  are the feature outputs for the $i$-th layer of the VGG-13 network. 

We also include a GAN losses that allow the network to generate realistic images.

The GAN loss $\mathcal{L}_\text{GAN}(G,D_1)$, using the global discriminator is given by
\small
\begin{multline}
\mathcal{L}_{\text{GAN}}(G,D_{1}) = \mathbb{E}_{(\vx, \vy_{\tgt})}[\log D_{1}(\vx, \vy_{\tgt}, \style_{\text{ref}})] \\ + \mathbb{E}[\log (1- D_{1}(G(\vx_{\text{ref}}, \vy_{\tgt}, \style_{\text{ref}}), \vy_{\tgt}, \style_{\text{ref}})].
\end{multline}
\normalsize
Further, the GAN loss $\mathcal{L}_\text{GAN}(G,D_2)$, using the patch discriminator is given by
\small
\begin{multline}
\mathcal{L}_{\text{GAN}}(G,D_{2}) = \mathbb{E}_{\vx}[\log D_{2}(\vx)] \\ + \mathbb{E}[\log (1- D_{2}(G(\vx_{\text{ref}}, \vy_{\tgt}, \style_{\text{ref}}))].
\end{multline}
\normalsize
We also include two perceptual cycle consistency losses:
\begin{equation}
\mathcal{L}_{\text{cyc}, \text{VGGFace}} = \mathbb{E}_{\vx}[\vggfaceloss(G(\vx^\prime, \vy_{\text{ref}}, \style^\prime), \vx)],
\end{equation}
\begin{equation}
\mathcal{L}_{\text{cyc}, \text{VGG}} = \mathbb{E}_{\vx}[\vggloss(G(\vx^\prime, \vy_{\text{ref}}, \style^\prime), \vx)],
\end{equation}
where $\vx^\prime = G(\vx, \vy_{\text{tgt}}, \style_{\text{ref}})$ is the generated sample and $\style^\prime = S(\vx^\prime)$ is the style vector predicted from this sample.

Finally, we have two supervised style losses to train the style encoder:
\begin{equation}
\mathcal{L}_{\text{sty}, \text{ref}} = \mathbb{E}_{\vx}[||\style^\prime - \style_{\text{ref}}||_1],
\end{equation}
\begin{equation}
\mathcal{L}_{\text{sty}, \text{tgt}} = \mathbb{E}_{\vx}[||\style^\prime - \style_{\text{tgt}}||_1].
\end{equation}
These two losses push the style encoder to generate the same style when the identity of the person is constant.

The final objective function is given by
\small
\begin{multline}
\mathcal{L} = \mathcal{L}_{\text{GAN}(G, D_1)} + \mathcal{L}_{\text{GAN}(G, D_2)} + \lambda_{\text{VGG}} \vggloss \\ + \lambda_{\text{VGGFace}} \vggfaceloss + \lambda_{\text{cyc}, \text{VGGFace}} \mathcal{L}_{\text{cyc}, \text{VGGFace}} \\ + \lambda_{\text{cyc}, \text{VGG}} \mathcal{L}_{\text{cyc}, \text{VGG}} + \lambda_{\text{sty}, \text{ref}} \mathcal{L}_{\text{sty}, \text{ref}} + \lambda_{\text{sty}, \text{tgt}} \mathcal{L}_{\text{sty}, \text{tgt}}.
\end{multline}
\normalsize

\subsection{Bias Testing using Face Synthesis}

We now describe bias detection in a trained face recognition network, $\Gamma$. To do this, we generate a dataset of synthetic samples $\mathcal{D}_\text{syn}$ of varying pose and expression for each identity in a set of real image samples $\mathcal{D}_\text{real}$. For each sample $\vx \in \mathcal{D}_\text{real}$, we fit the pose and expression parameters $\pose$ and $\expression$. We then generate new images $\vx_\text{syn} = G(\vx, \pose_\text{syn}, \expression_\text{syn}) $ by sampling a single dimension of $\pose$ or $\expression$ in a grid-like fashion within a pre-set range such that $\pose_\text{syn} = (1+k) \cdot \pose$ and $\expression_\text{syn} = (1+k) \cdot \expression$ with $k \in (-K, K)$ and $K$ is a constant.  Finally, we render the resulting images using MorphGAN.

Once we have the generated dataset for a specific attribute change, we use our face recognition network to recognize the faces in the generated dataset and plot the response curves to see the degradation in performance as the attribute changes. The generated synthetic images can also be used to augment the real training dataset to alleviate identified biases.  For example, if the recognition network performs poorly at a given pose angle, then more samples can be generated at that particular pose angle for training.

\section{Experiments}\label{sec:experiments}

\begin{figure*}[t]
    \includegraphics[width=\textwidth]{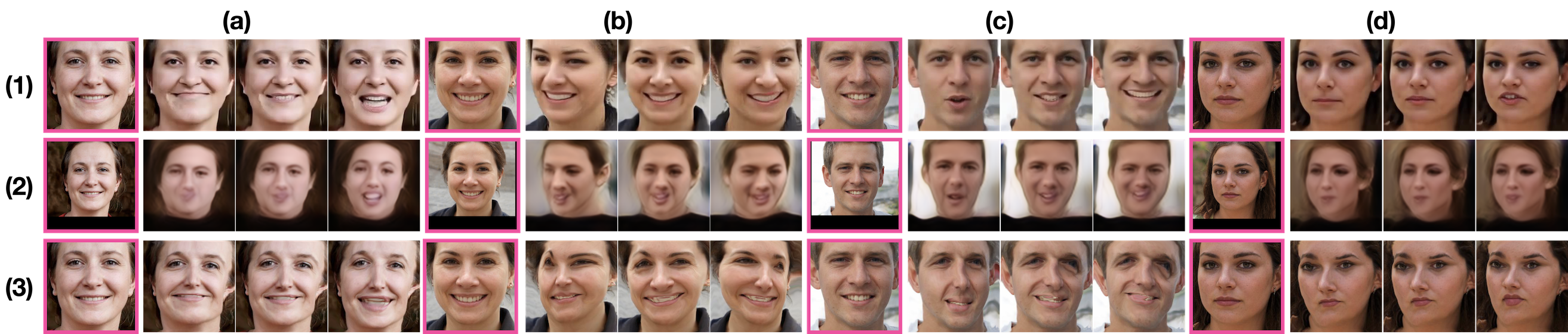}
\caption{Comparison between (1) our face synthesis network MorphGAN, (2) an unofficial open-source implementation of Zakharov et al.~\cite{zakharov2019few} and (3) the official implementation of Wiles et al.~\cite{wiles2018x2face}. 
The presented expression and pose changes are \emph{(a)} jaw close/open, \emph{(b)} yaw rotation, \emph{(c)} smile and \emph{(d)} pucker.  For each example, the pink highlighted image is the reference image.
}
\label{fig:comparison_experiment}
\end{figure*}

\begin{table*}[t]
\begin{tabular}{lcccccc}
\Xhline{3\arrayrulewidth}
& \multicolumn{2}{c}{\bf One Sample} & \multicolumn{2}{c}{\bf Two Samples} & \multicolumn{2}{c}{\bf Three Samples} \\
{\bf Parameter} & Normal & Augmented & Normal & Augmented & Normal & Augmented \\ \hline
Jaw & 71.0\% & \textbf{77.0\%} & 86.2\% & \textbf{89.0\%} & 90.3\% & \textbf{90.5\%} \\ 
Yaw & 75.8\% & \textbf{79.9\%} & 88.5\% & \textbf{90.7\%} & \textbf{91.7\%} & 91.4\% \\ 
Smile & 69.5\% & \textbf{76.4\%} & 86.1\% & \textbf{88.2\%} & 89.0\% & \textbf{90.2\%} \\ 
Pucker & 68.9\% & \textbf{77.7\%} & 88.1\% & \textbf{88.3\%} & 90.4\% & \textbf{91.7\%} \\
Eyebrow & 73.4\% & \textbf{73.5\%} & 86.6\% & \textbf{86.7\%} & 89.9\% & \textbf{91.2\%} \\ \Xhline{3\arrayrulewidth}
\end{tabular}
\centering
\caption{Face recognition results for a Inception ResNet v1 network, pre-trained on VGGFace2 and fine-tuned on 150 identities from the AVSpeech Test dataset. We display performance for a normally trained network (normal), and a network that is pre-trained on synthetically augmented samples generated by MorphGAN (aug). The top row designates how many real training samples are used to train the network.}
\label{table:augmentation_experiment}
\end{table*}

We present three sets of experiments -- \emph{diagnosis experiments}, \emph{augmentation experiments} and \emph{face synthesis}. In the diagnosis experiments, we test the sensitivity of the face recognition model using controlled trials where identity is fixed and either pose or a specific expression parameter is varied.  We plot performance curves for the respective pose and expression parameters under test to show the impact on the model. In the augmentation experiments, we show that a facial recognition network trained by augmenting a face dataset using MorphGAN performs better when training data are limited. Moreover, augmentation of specific expression and pose changes that result in robustness issues identified in the diagnosis experiments leads to improved performance. Finally, we show qualitative examples of our face synthesis using MorphGAN on samples from sources other than our training dataset (AVSpeech). We also compare our results with related work.

To generate face shape renders, we use the FLAME face model~\cite{flame}, parameterized by shape, expression and pose parameters. Our generative face model, MorphGAN, is trained on a subset of the AVSpeech dataset~\cite{ephrat2018looking}.  AVSpeech contains thousands of hours of video segments from various sources.  We select a subset of 13,000 videos from this dataset for training, and we sample a random subset of 150 videos of different people from the test set for testing.  We detect the face in each frame and extract the 2D facial landmarks using~\cite{king2009dlib,bulat2017far}, fit FLAME parameters using the landmarks, and render the resulting shape with a gray texture (see Figure~\ref{fig:diagnosis_experiment} for example renders).

In all experiments, our face recognition model is an Inception-ResNet-v1 network~\cite{szegedy2016inception} pre-trained on the VGGFace2 dataset~\cite{Cao18} using the FaceNet triplet loss~\cite{schroff2015facenet}.

\subsection{Diagnosis Experiments}\label{sec:experiments-diagnosis}
Diagnosis experiments discover robustness issues in a trained facial recognition network using face images generated by MorphGAN by systematically varying one expression parameter or one pose parameter. We evaluate the performance of the face recognition model as a function of the change in the given parameter.

We fine-tune the pre-trained recognition network using two frames for each of the 150 identities from the AVSpeech test set.  Next, using MorphGAN we generate new frames by varying the expression or pose for the following parameters: yaw rotation, jaw opening, smile, lip pucker and eyebrow raise. In total, we generate 21 synthetic frames for each of the selected parameters, and for each of the images used for fine-tuning.

We plot the recognition accuracy for the generated images in Figure~\ref{fig:diagnosis_experiment}. Each accuracy curve has the ground-truth pose and expression parameters at the center of the x-axis, and varies in the negative and positive direction towards the left and right on the x-axis, respectively. Underneath each curve, we present illustrative examples of the generated faces and the FLAME model render used to generate each face. 

In most cases, the highest accuracy is obtained for the ground-truth pose and expression parameter value. The exceptions are for the smile and pucker parameters, which have a higher peak for a slightly modified expression in the positive direction. In all of the examples in Figure~\ref{fig:diagnosis_experiment}, we can see that the recognition accuracy drops as the expression and pose become increasingly different from the ground-truth. We observe that changing the yaw rotation parameter has a large negative impact on accuracy (decrease from $>80\%$ to $\approx 46\%$). Varying the degree of mouth opening, lip pucker and the smile parameters also degrades performance, but less drastically (accuracy decreases by up to $7\%$).  In contrast, modifying the eyebrow lower/raise parameter has only a modest impact on accuracy (decrease by $\approx 2\%$) because the resultant change in the image from this parameter is smaller than the change that results from the other parameters.  We also observe that the response curve for some parameters is (approximately) symmetric for positive and negative changes in the parameter value. For example, by varying the yaw in the positive and negative directions by the same magnitude results in a similar degradation in recognition accuracy. By contrast, we observe asymmetric responses in the jaw close/open and smile curves, where the accuracy is less affected when the parameter is changed in the positive direction.  This asymmetry could result from bias due to seeing fewer samples at one end of the spectrum of the particular parameter during training than at the other end of the spectrum.

We observe a strong dependency of the facial recognition network with respect to large changes that occlude face regions, such as changes in yaw. This dependency is similar to findings using real world data. We hypothesize that robustness issues with respect to each of the pose and expression changes that we have probed are a result of bias due to limited training data. Some of these robustness issues are asymmetric, and the impact on accuracy is less severe for parameters that result in less change in the image, such as eyebrow raise/lower.

\subsection{Augmentation Experiments}

In the following experiments, we augment the training data with images generated using MorphGAN and fine-tune the face recognition model to show improvement in recognition performance in a small data scenario.

The diagnosis experiments in Section \ref{sec:experiments-diagnosis} provided a way to use synthetic samples from MorphGAN to probe the impact on model performance. However, there potentially is a domain gap between synthetic data and real data.  Therefore, to validate our diagnosis, we augment the training datasets using synthetic images from MorphGAN with varying pose and expressions. As in Section \ref{sec:experiments-diagnosis}, we vary parameters independently to measure their specific effect on recognition accuracy. If a certain augmentation in expression or pose improves results over the original model, it supports the hypothesis that the original model is not robust to such a variation.

We use the pretrained Inception-ResNet-v1 network as the face recognition model. We then fine-tune this network under two cases:  firstly using only real samples from the AVSpeech dataset, and secondly by augmenting these samples using MorphGAN to get more variation in head pose and expression. We compare the recognition accuracy in both cases for changes in the same five parameters described in Section \ref{sec:experiments-diagnosis}. We also compare the performance when the number of real training samples varies from one to three per identity.

To generate our training and test datasets for the facial recognition task, we fit the FLAME model to all frames in the 150 selected AVSpeech test set videos. Next, to build an \emph{unbiased}, uniformly sampled test set and a \emph{biased} training set, we generate test and training sets for each parameter individually. For the training set, we sample frames that are closest to the mean parameter value of each video, and for the test set we sample 10 frames by uniformly sampling the chosen parameter. In this way, we build an unbiased uniformly sampled test set and a training set that is biased towards the mean for each parameter. Both of these sets are composed of only real samples.

The results for these experiments are shown in Table~\ref{table:augmentation_experiment}. We observe that in the small-data scenario, where only one training sample per identity is available, we obtain the most gain in performance by augmenting the data set using images from MorphGAN for each pose and expression parameter, except eyebrow lower/raise expression. This increase in performance is in line with the conclusions from our diagnosis experiment in Section \ref{sec:experiments-diagnosis}, where the impact on accuracy for eyebrow lower/raise was minimal, but was more significant for the other parameters. Overall, we observe increases in accuracy after augmenting the training data with synthetic samples. We also observe that the gains in accuracy reduce as more real data are available. The benefits almost saturate once we have at least three real samples per identity.

These results show that MorphGAN can be an effective way of augmenting a dataset for facial recognition, especially in scenarios where there is a limited number of images for each identity, and in which the training data are biased and do not capture the variance with respect to a specific pose or expression change. This experiment also validates the fact that MorphGAN preserves identity while successfully varying pose and expression, since it improves facial recognition accuracy. Finally, we see that we obtain the most significant performance increases (up to $9\%$) when we augment our training dataset using expression and pose changes that most impacted the facial recognition network in the diagnosis experiment.

\subsection{Qualitative Samples}

\begin{figure*}[t]
    \centering
    \includegraphics[width=0.90\textwidth]{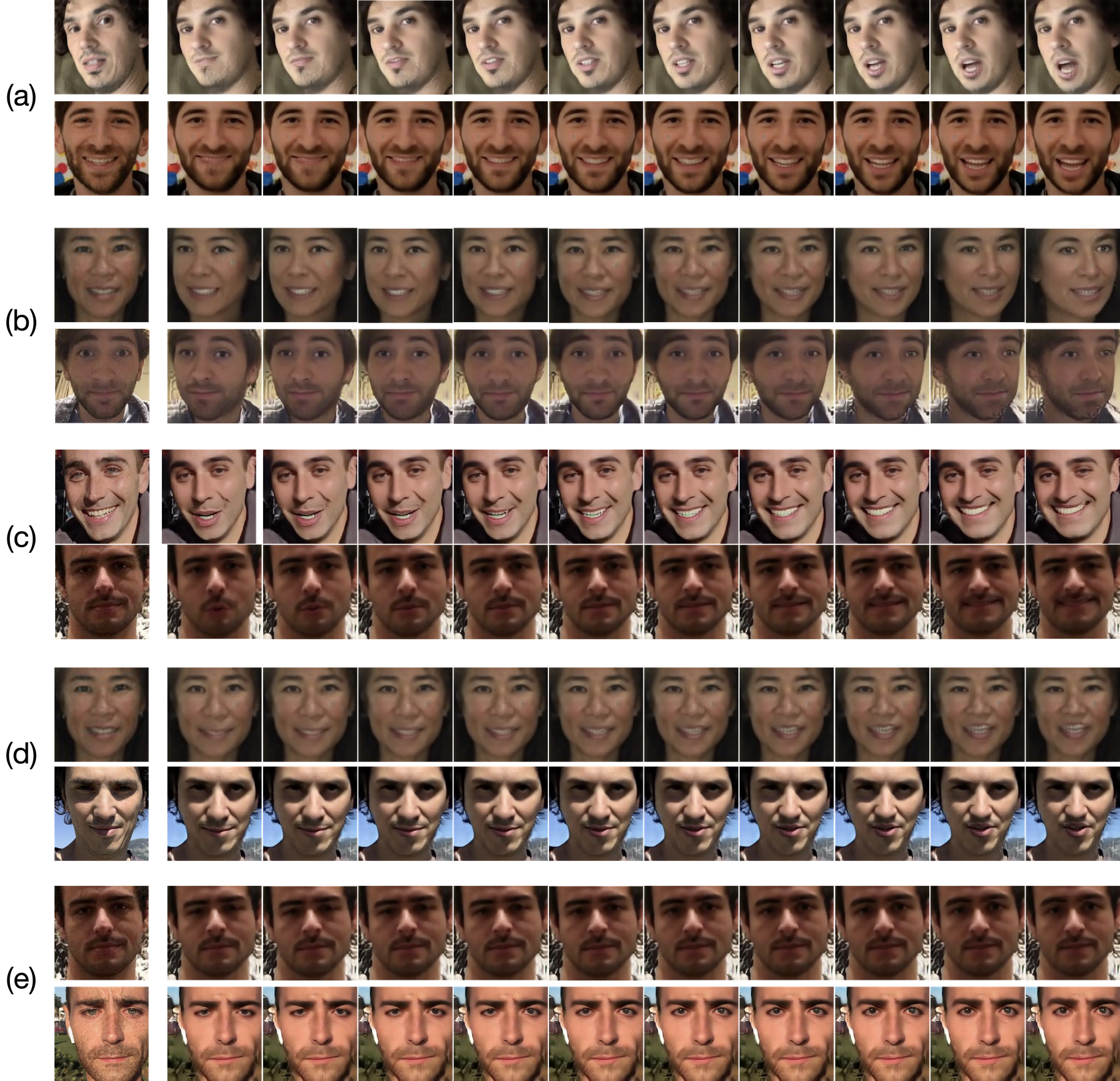}
    \caption{Examples of facial expression and head pose changes using MorphGAN on example faces of real people that expressively granted permission for use as shown in this work. The presented expression and pose changes are \emph{(a)} jaw close/open, \emph{(b)} yaw rotation, \emph{(c)} smile, \emph{(d)} pucker and \emph{(e)} eyebrow lower/raise.}
    \label{fig:qualitative_air}
\end{figure*}

\begin{figure*}[t]
    \centering
    \includegraphics[width=0.86\textwidth]{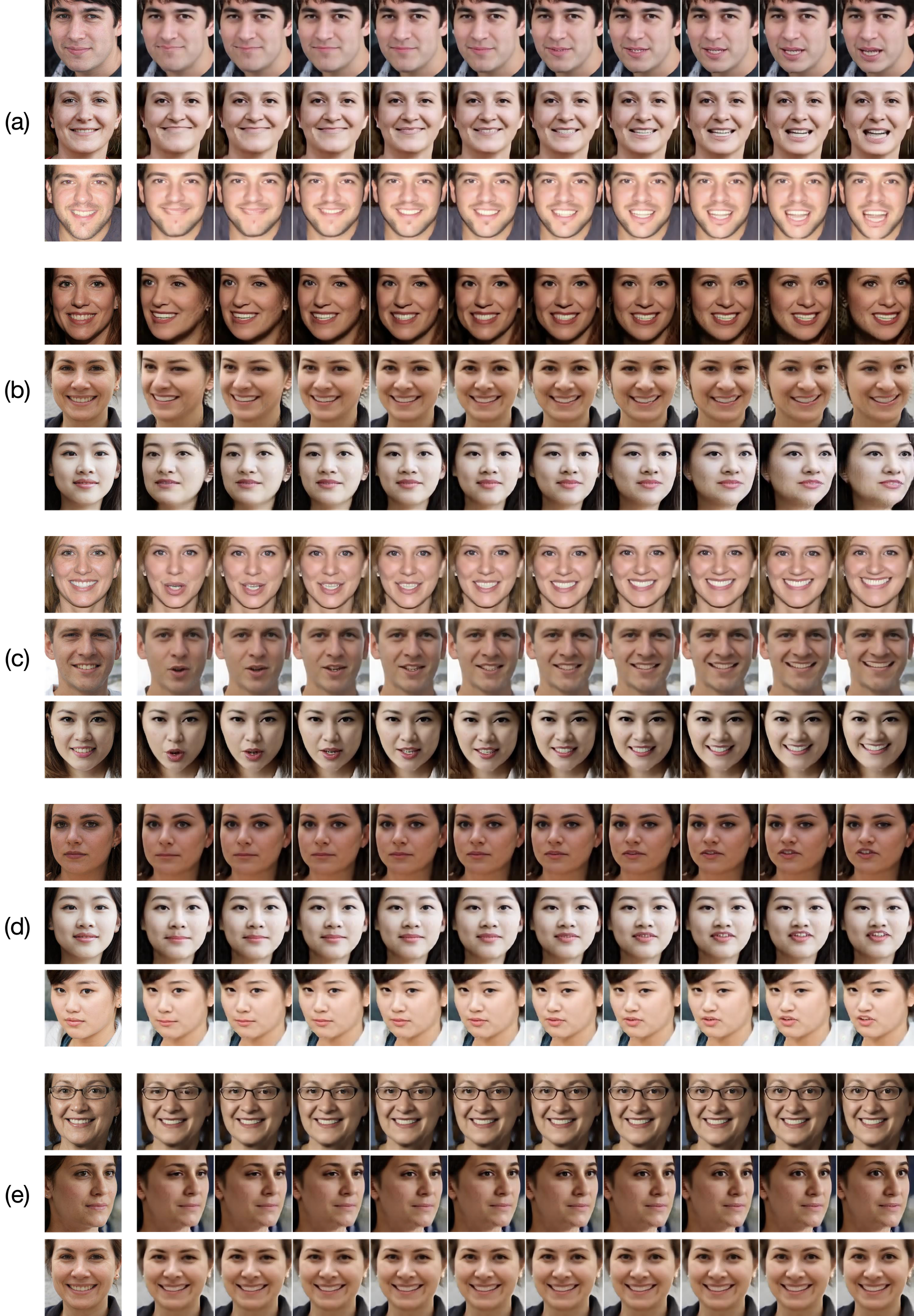}
    \caption{Examples of facial expression and head pose changes using MorphGAN on example face images generated using the StyleGAN2 architecture trained on the FFHQ dataset. The presented expression and pose changes are \emph{(a)} jaw close/open, \emph{(b)} yaw rotation, \emph{(c)} smile, \emph{(d)} pucker and \emph{(e)} eyebrow lower/raise.}
    \label{fig:qualitative_stylegan}
\end{figure*}

We present qualitative samples of our one-shot face synthesis model to validate (1) identity preservation, (2) expression fidelity and variance, and (3) pose fidelity. In Figure~\ref{fig:comparison_experiment}, we show comparisons between MorphGAN, an unofficial open-source implementation of Zakharov et al.~\cite{zakharov2019few} with pre-trained weights, and the official implementation of X2Face~\cite{wiles2018x2face}. Input images for Zakharov et al.~\cite{zakharov2019few} are zoomed out and padded in order to be inside of their training domain. In Figure~\ref{fig:qualitative_air}, we show qualitative samples on \textbf{real faces} for individuals that have given express permission to include their images in this paper. In Figure~\ref{fig:qualitative_stylegan}, we show examples of expression and pose changes on \textbf{synthetic faces} generated using the StyleGAN2 architecture trained on the FFHQ dataset.

\section{Conclusion}\label{sec:conclusion}
In this paper we have presented one-shot generation of images of people with novel facial expressions and head poses using a GAN with interpretable controls.  We have used the network to show how bias can be detected in a trained facial recognition network.  To synthesize faces we adopt a two step approach:  firstly, render the face shape with the desired expression and pose, and secondly render the final image.  In particular, we introduce an image generator conditioned on a reference image and the target shape render.

We show that our synthesized face images preserve the identity in the original image, and the synthesized images have high fidelity in expression and pose changes. We have also shown that we can diagnose potential sources of bias with respect to pose and expression.  Finally, we have shown that we can improve facial recognition results in a small-data environment by augmenting a source training dataset using additional synthesized face images with new expressions and poses for the corresponding existing identities the training set.

{\bf Privacy and Consent.}
The people in the real samples shown in this paper (Figure~\ref{fig:qualitative_air}) explicitly granted consent for their use in this work. The example synthetic faces (Figure~\ref{fig:qualitative_stylegan}) are generated using GAN and thus the faces do not correspond to real people.

{\bf Acknowledgements.}
We are grateful to Russ Webb and Ashish Strivastava for useful feedback.

\section*{A. Appendix}
\subsection*{A.1. Network Architecture}

We use a similar architecture to the generator and style encoder in StarGAN-v2~\cite{choi2020stargan}, but modify it to incorporate our shape renders. We design our own global discriminator $D_1$ and patch discriminator $D_2$, as described in the following sections.

\paragraph{Generator}
We use a generator with four down-sampling blocks, four residual blocks and four up-sampling blocks as used by StarGAN-v2~\cite{choi2020stargan}. All blocks have residual connections. Instance Normalization (IN)~\cite{ulyanov2016instance} is used for the down-sampling blocks as well as the first two residual blocks. Adaptive Instance Normalization (AdaIN)~\cite{huang2017arbitrary} blocks are used for the last two residual blocks and for the up-sampling blocks. The AdaIN layers take a style encoding as input. 
The main difference for our generator is that it takes the reference image and the concatenated target shape render as input, in the form of a 256x256x6 tensor.

\paragraph{Style Encoder}
We use a similar style encoder to StarGAN v2~\cite{choi2020stargan}, except we only use one output branch.  The input to the network is a face image.  The network itself is composed of an initial 1x1 convolutional layer, followed by six down-sampling residual blocks, a Leaky ReLU, a 4x4 convolutional layer that maps the feature map to a 1x1x512 vector, another Leaky ReLU and a final linear layer.

\paragraph{Global Discriminator}

We use two branches -- an image/render branch and a style branch. The image/render branch has six pre-activation residual blocks with leaky ReLUs using the same layout as the style encoder. The style branch has four linear layers [[1024, 512], [512, 256], [256, 128], and [128,32]] all using ReLU activations. Finally, after concatenation of the output features from the image/render branch and the style branch, there is a final branch with three linear layers [[64, 32], [32, 16], and [16, 1]], mapping to a final prediction. The Global discriminator takes in a face image (real or fake), a shape render and a style vector as input.

\paragraph{Patch Discriminator}

Our patch discriminator only takes in the face image (real or fake) as an input. We use four pre-activation residual blocks in the same fashion as the image/render branch of our global discriminator. The final patch sizes are 8x8.

\subsection*{A.2. Training Details}

We use the following values for the full loss presented in the main paper: $\lambda_{\text{VGG}} = 10, \lambda_{\text{VGGFace}} = 10, \lambda_{\text{cyc}, \text{VGGFace}} = 1, \lambda_{\text{cyc}, \text{VGG}} = 1, \lambda_{\text{sty}, \text{ref}} = 1, \lambda_{\text{sty}, \text{tgt}} = 1$.

We train the network for 30 epochs on our dataset of 13,000 videos (corresponding to approximately 1M image pairs). We use a batch size of 128, and  Adam~\cite{kingma2014adam} optimizers with a learning rate of $1e-4$, weight decay $1e-4$, $\beta_1 = 0.0$ and $\beta_2 = 0.99$.

\paragraph{Balancing Pose in Batches}
We pre-compute pose differences between image pairs in videos, and bin pose differences between the reference image and the target image into the following bins: [[-180 \textdegree, -30 \textdegree], [-30 \textdegree, -15 \textdegree], [-15 \textdegree, -5 \textdegree], [-5 \textdegree, 5 \textdegree], [5 \textdegree, 30 \textdegree], [30 \textdegree, 180 \textdegree]]. We then balance our batches with an equal number of samples from each bin to train with variation in pose differences between reference and target images. This improves the pose fidelity of MorphGAN as well as alleviate artifacts in the rendered output when pose changes are large.

\subsection*{A.3. Qualitative Results}

\begin{figure*}[t]
    \centering
    \includegraphics[width=0.6\textwidth]{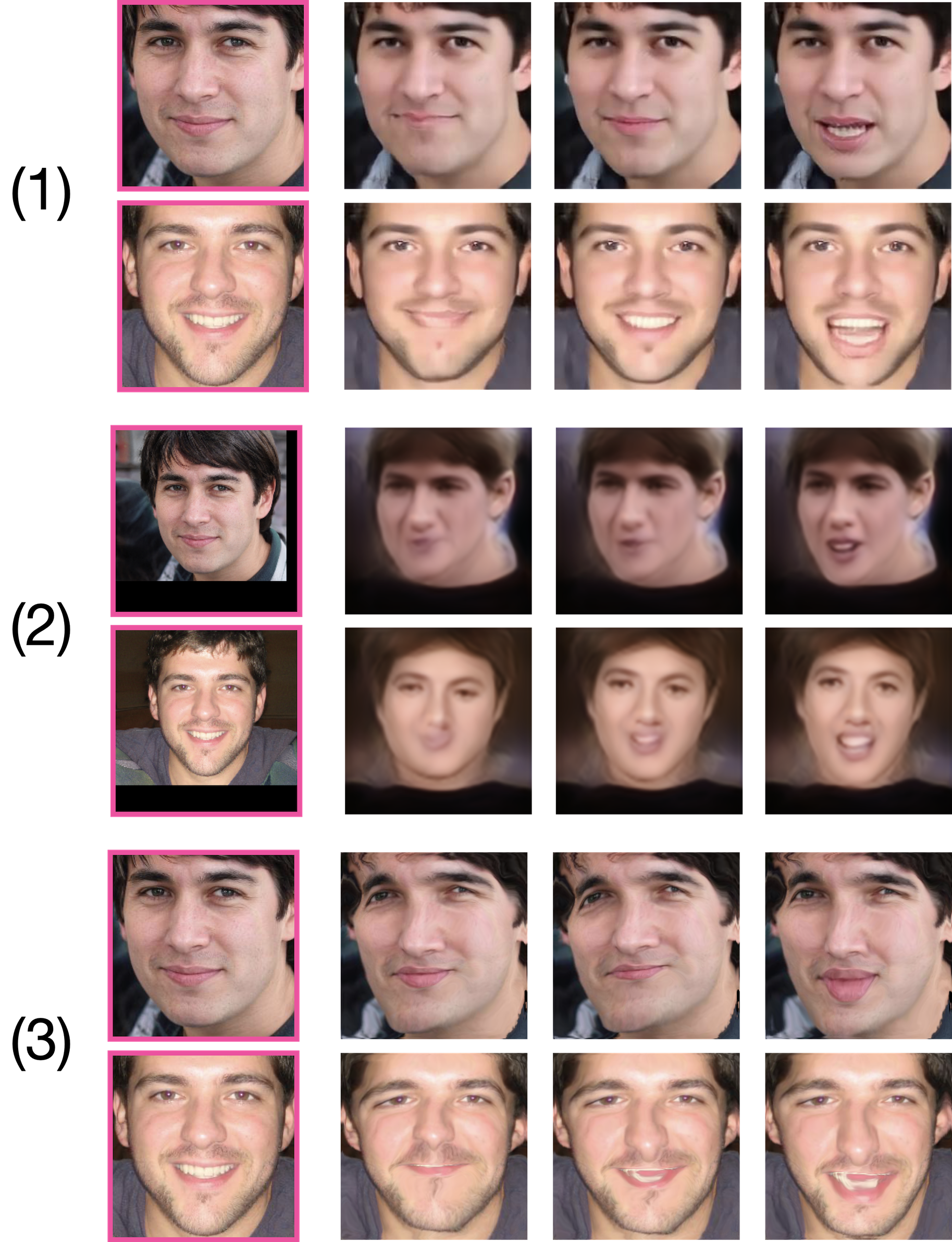}
    \caption{Comparison between (1) our face synthesis network MorphGAN, (2) an unofficial open-source implementation of Zakharov et al.~\cite{zakharov2019few} and (3) the official implementation of Wiles et al.~\cite{wiles2018x2face}. 
    The presented expression and pose changes are for ``jaw close/open'' sequence. For each example, The pink highlighted image is the reference image.}
    \label{fig:comparison_supp_1}
\end{figure*}

\begin{figure*}[t]
    \centering
    \includegraphics[width=0.6\textwidth]{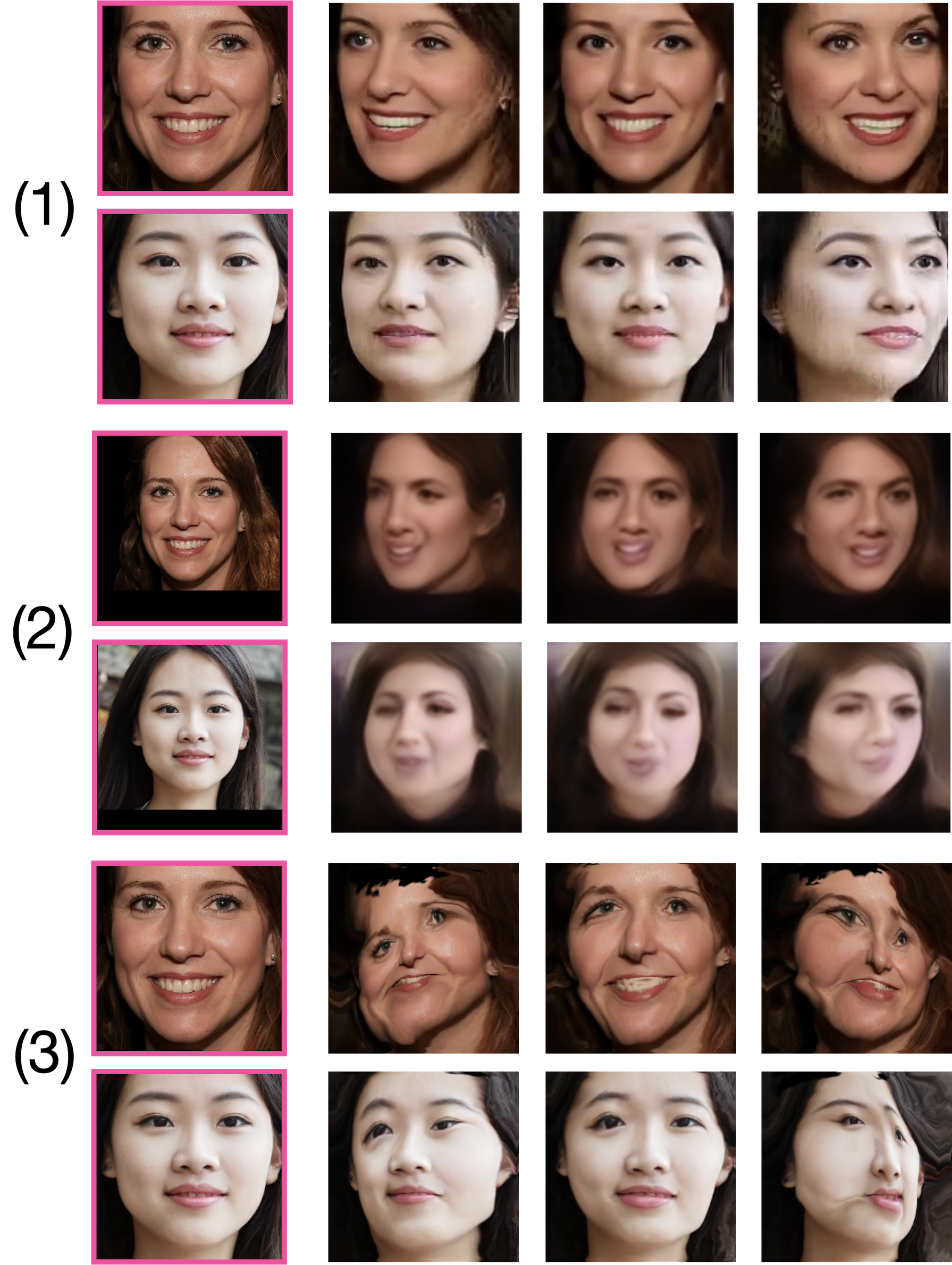}
    \caption{Comparison between (1) our face synthesis network MorphGAN, (2) an unofficial open-source implementation of Zakharov et al.~\cite{zakharov2019few} and (3) the official implementation of Wiles et al.~\cite{wiles2018x2face}. 
    The presented expression and pose changes are for ``yaw rotation'' sequence. For each example, The pink highlighted image is the reference image.}
    \label{fig:comparison_supp_2}
\end{figure*}

\begin{figure*}[t]
    \centering
    \includegraphics[width=0.6\textwidth]{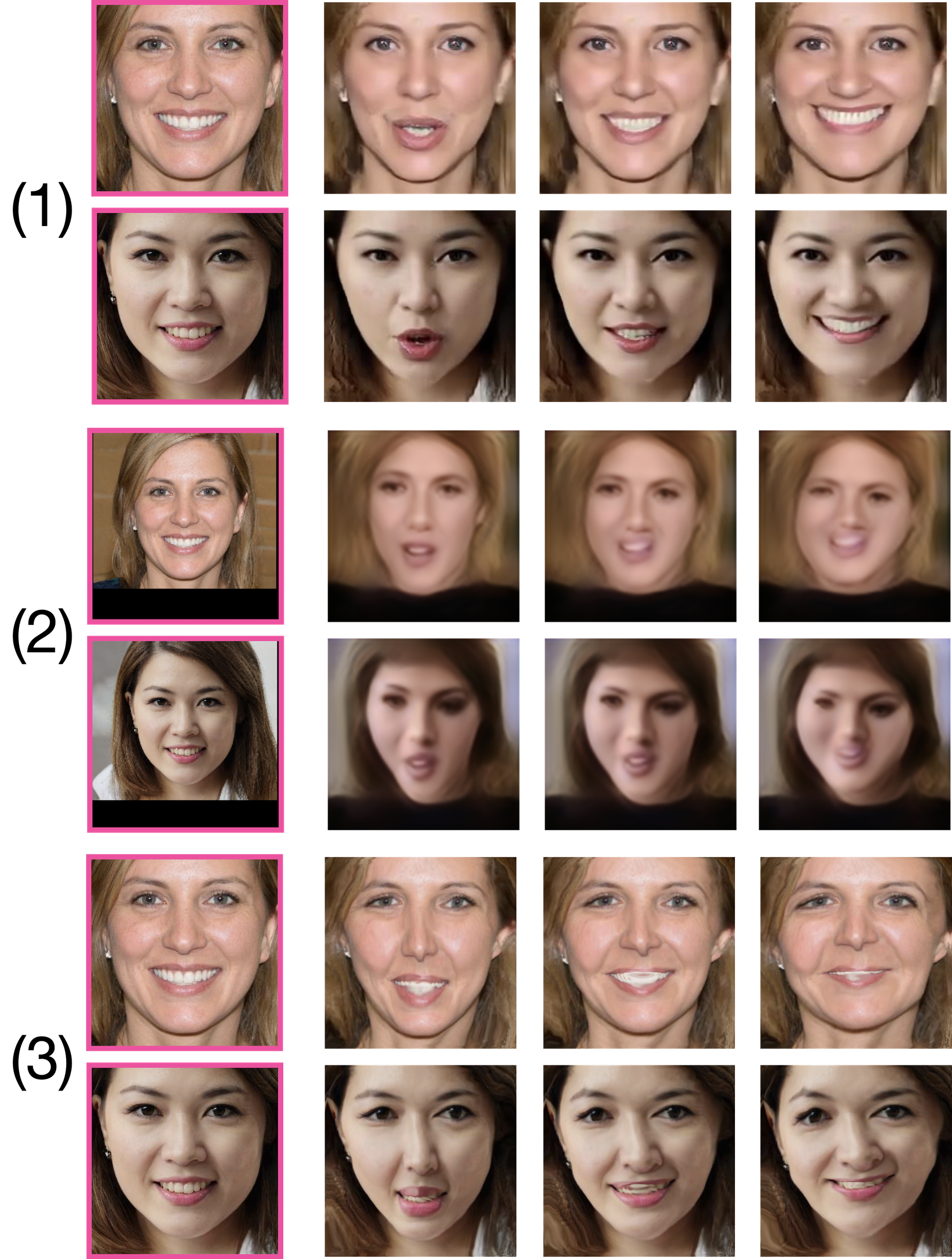}
    \caption{Comparison between (1) our face synthesis network MorphGAN, (2) an unofficial open-source implementation of Zakharov et al.~\cite{zakharov2019few} and (3) the official implementation of Wiles et al.~\cite{wiles2018x2face}. 
    The presented expression and pose changes are for ``smile'' sequence.  For each example, the pink highlighted image is the reference image.}
    \label{fig:comparison_supp_3}
\end{figure*}

\begin{figure*}[t]
    \centering
    \includegraphics[width=0.6\textwidth]{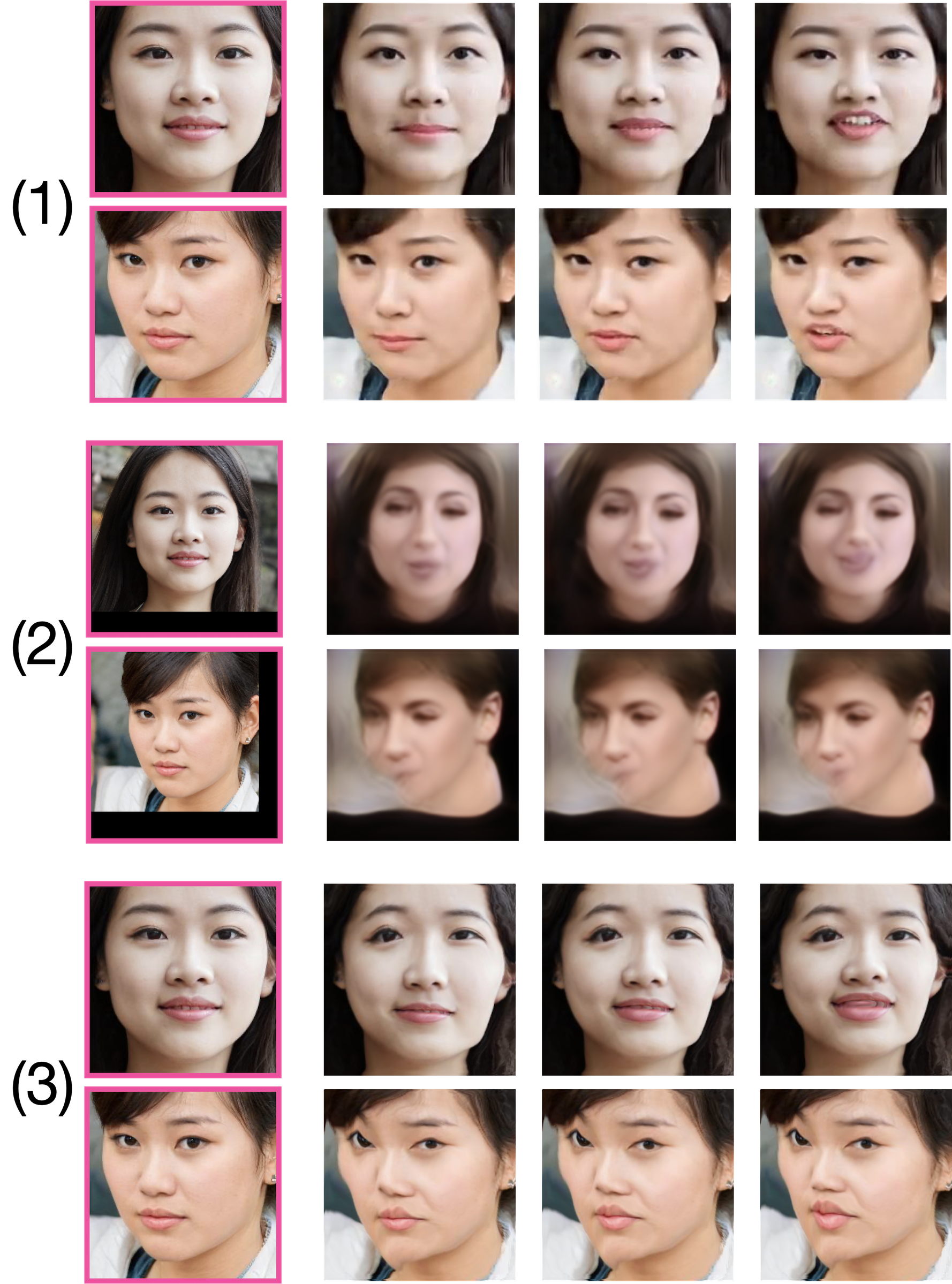}
    \caption{Comparison between (1) our face synthesis network MorphGAN, (2) an unofficial open-source implementation of Zakharov et al.~\cite{zakharov2019few} and (3) the official implementation of Wiles et al.~\cite{wiles2018x2face}. 
    The presented expression and pose changes are for ``pucker'' sequence.  For each example, the pink highlighted image is the reference image.}
    \label{fig:comparison_supp_4}
\end{figure*}

\begin{figure*}[t]
    \centering
    \includegraphics[width=0.6\textwidth]{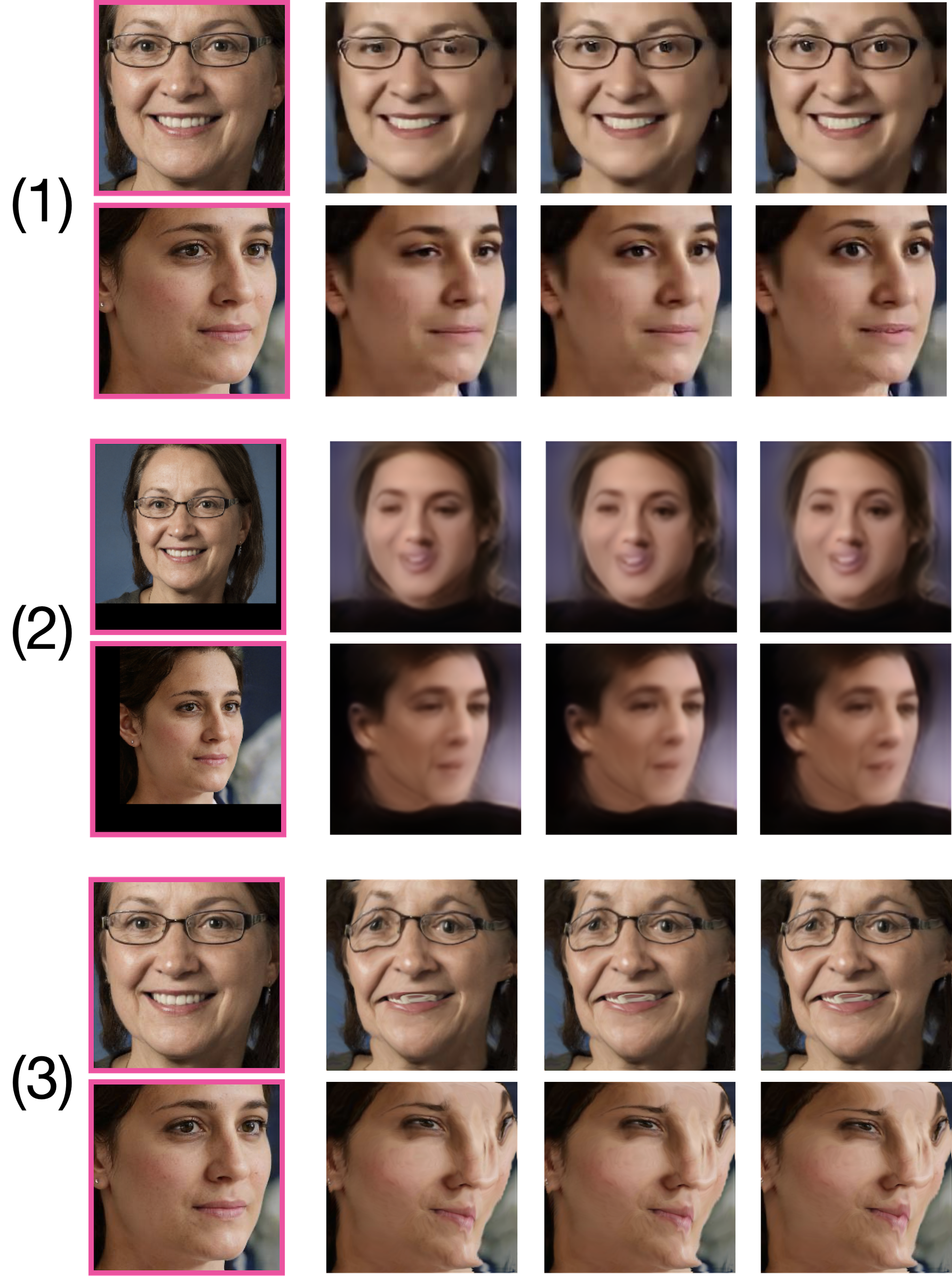}
    \caption{Comparison between (1) our face synthesis network MorphGAN, (2) an unofficial open-source implementation of Zakharov et al.~\cite{zakharov2019few} and (3) the official implementation of Wiles et al.~\cite{wiles2018x2face}. 
    The presented expression and pose changes are for ``eyebrow lower/raise'' sequence.  For each example, the pink highlighted image is the reference image.}
    \label{fig:comparison_supp_5}
\end{figure*}

We show additional comparative qualitative results of our one-shot face synthesis model. In Figures~\ref{fig:comparison_supp_1}, \ref{fig:comparison_supp_2}, \ref{fig:comparison_supp_3}, \ref{fig:comparison_supp_4}, \ref{fig:comparison_supp_5}, we show comparisons between MorphGAN, an unofficial open-source implementation of Zakharov et al.~\cite{zakharov2019few} with pre-trained weights, and the official implementation of X2Face~\cite{wiles2018x2face}. Input images for Zakharov et al.~\cite{zakharov2019few} are zoomed out and padded to be within their training domain. We show five different expression and pose changes: jaw close/open, yaw rotation, smile, pucker and eyebrow lower/raise.

\subsection*{A.4. Comparison with Zakharov et al.~\cite{zakharov2019few}}

All the samples generated using Zakharav et al.~\cite{zakharov2019few} use their generator in one-shot mode for a direct comparison with our method. It should be noted that most of the visual results in Zakharav et al.~\cite{zakharov2019few} are generated using few-shot mode, where multiple samples of the source identity are available, which leads to a better quality of generated images. Due to lack of official source code, we use an open-source version of the code~\footnote{github.com/vincent-thevenin/Realistic-Neural-Talking-Head-Models} that has been trained for five epochs in a smaller version of the full VoxCeleb2~\cite{Chung18b}.

\clearpage

{\small
\bibliographystyle{ieee_fullname}
\bibliography{egbib}

\begin{thebibliography}{10}\itemsep=-1pt

\bibitem{abate20072d}
A. Abate, M. Nappi, D. Riccio, and G. Sabatino.
\newblock 2d and 3d face recognition: A survey.
\newblock {\em Pattern recognition letters}, 28(14):1885--1906, 2007.

\bibitem{Alvi_2018_ECCV_Workshops}
M. Alvi, A. Zisserman, and C. Nellaaker.
\newblock Turning a blind eye: Explicit removal of biases and variation from
  deep neural network embeddings.
\newblock In {\em Proceedings of the European Conference on Computer Vision
  (ECCV) Workshops}, September 2018.

\bibitem{bellamy2019ai}
R Bellamy, K Dey, M Hind, S Hoffman, S. Houde, K. Kannan, P. Lohia, J. Martino,
  S. Mehta, A. Mojsilovi{\'c}, S. Nagar, K. Natesan~Ramamurthy, J. Richards, D.
  Saha, P. Sattigeri, M. Singh, K. Varshney, and Y. Zhang.
\newblock Ai fairness 360: An extensible toolkit for detecting and mitigating
  algorithmic bias.
\newblock {\em IBM Journal of Research and Development}, 63(4/5):4--15, 2019.

\bibitem{blanz1999morphable}
V. Blanz and T. Vetter.
\newblock A morphable model for the synthesis of 3d faces.
\newblock In {\em SIGGRAPH}, volume~99, pages 187--194, 1999.

\bibitem{bulat2017far}
A. Bulat and G. Tzimiropoulos.
\newblock How far are we from solving the 2d \& 3d face alignment problem?(and
  a dataset of 230,000 3d facial landmarks).
\newblock In {\em Proceedings of the IEEE International Conference on Computer
  Vision}, pages 1021--1030, 2017.

\bibitem{buolamwini2018gender}
Joy Buolamwini and Timnit Gebru.
\newblock Gender shades: Intersectional accuracy disparities in commercial
  gender classification.
\newblock In {\em Conference on fairness, accountability and transparency},
  pages 77--91, 2018.

\bibitem{cao2018vggface2}
Qiong Cao, Li Shen, Weidi Xie, Omkar~M Parkhi, and Andrew Zisserman.
\newblock Vggface2: A dataset for recognising faces across pose and age.
\newblock In {\em 2018 13th IEEE International Conference on Automatic Face \&
  Gesture Recognition (FG 2018)}, pages 67--74. IEEE, 2018.

\bibitem{Cao18}
Q. Cao, L. Shen, W. Xie, O.~M. Parkhi, and A. Zisserman.
\newblock Vggface2: A dataset for recognising faces across pose and age.
\newblock In {\em International Conference on Automatic Face and Gesture
  Recognition}, 2018.

\bibitem{cavazos2020accuracy}
Jacqueline~G Cavazos, P~Jonathon Phillips, Carlos~D Castillo, and Alice~J
  O’Toole.
\newblock Accuracy comparison across face recognition algorithms: Where are we
  on measuring race bias?
\newblock {\em IEEE Transactions on Biometrics, Behavior, and Identity
  Science}, 2020.

\bibitem{Cavazos_2020}
J.~G. {Cavazos}, P.~J. {Phillips}, C.~D. {Castillo}, and A.~J. {O’Toole}.
\newblock Accuracy comparison across face recognition algorithms: Where are we
  on measuring race bias?
\newblock {\em IEEE Transactions on Biometrics, Behavior, and Identity
  Science}, 2020.

\bibitem{chen:2019:hierarchical}
L. Chen, R. Maddox, Z. Duan, and C. Xu.
\newblock Hierarchical cross-modal talking face generation with dynamic
  pixel-wise loss.
\newblock In {\em Proceedings of the IEEE Conference on Computer Vision and
  Pattern Recognition}, pages 7832--7841, 2019.

\bibitem{chen:2019:photo-realistic}
Z. Chen, G. Zhang, Z. Zhang, K. Mitchell, and J. Yu.
\newblock Photo-realistic facial details synthesis from single image.
\newblock {\em arXiv preprint arXiv:1903.10873}, 2019.

\bibitem{choe2017face}
J. Choe, S. Park, K. Kim, J. Hyun~Park, D. Kim, and H. Shim.
\newblock Face generation for low-shot learning using generative adversarial
  networks.
\newblock In {\em Proceedings of the IEEE International Conference on Computer
  Vision Workshops}, pages 1940--1948, 2017.

\bibitem{choi2020stargan}
Yunjey Choi, Youngjung Uh, Jaejun Yoo, and Jung-Woo Ha.
\newblock Stargan v2: Diverse image synthesis for multiple domains.
\newblock In {\em Proceedings of the IEEE/CVF Conference on Computer Vision and
  Pattern Recognition}, pages 8188--8197, 2020.

\bibitem{Chung18b}
J.~S. Chung, A. Nagrani, and A. Zisserman.
\newblock Voxceleb2: Deep speaker recognition.
\newblock In {\em INTERSPEECH}, 2018.

\bibitem{history3dmm}
B. Egger, William~AP Smith, Ayush Tewari, Stefanie Wuhrer, Michael Zollhoefer,
  Thabo Beeler, Florian Bernard, Timo Bolkart, Adam Kortylewski, Sami Romdhani,
  et~al.
\newblock 3d morphable face models—past, present, and future.
\newblock {\em ACM Transactions on Graphics (TOG)}, 39(5):1--38, 2020.

\bibitem{ephrat2018looking}
A. Ephrat, I. Mosseri, O. Lang, T. Dekel, K. Wilson, A. Hassidim, W.~T Freeman,
  and M. Rubinstein.
\newblock Looking to listen at the cocktail party: a speaker-independent
  audio-visual model for speech separation.
\newblock {\em ACM Transactions on Graphics (TOG)}, 37(4):112, 2018.

\bibitem{Garcia_2019}
R.~V. {Garcia}, L. {Wandzik}, L. {Grabner}, and J. {Krueger}.
\newblock The harms of demographic bias in deep face recognition research.
\newblock In {\em 2019 International Conference on Biometrics (ICB)}, pages
  1--6, 2019.

\bibitem{gatys2015artistic}
L. Gatys, A. Ecker, and M. Bethge.
\newblock A neural algorithm of artistic style.
\newblock In {\em CoRR}, 2015.

\bibitem{gatys2015texture}
L. Gatys, A. Ecker, and M. Bethge.
\newblock Texture synthesis using convolutional neural networks.
\newblock In {\em Advances in Neural Information Processing Systems}, pages
  262--270, 2015.

\bibitem{geng20193d}
Z. Geng, C. Cao, and S. Tulyakov.
\newblock 3{D} guided fine-grained face manipulation.
\newblock In {\em Proceedings of the IEEE Conference on Computer Vision and
  Pattern Recognition}, pages 9821--9830, 2019.

\bibitem{ghosh2020gif}
Partha Ghosh, Pravir~Singh Gupta, Roy Uziel, Anurag Ranjan, Michael Black, and
  Timo Bolkart.
\newblock Gif: Generative interpretable faces.
\newblock 2019.

\bibitem{goodfellow2014generative}
I. Goodfellow, J. Pouget-Abadie, M. Mirza, B. Xu, D. Warde-Farley, S. Ozair, A.
  Courville, and Y. Bengio.
\newblock Generative adversarial nets.
\newblock In {\em Advances in neural information processing systems}, pages
  2672--2680, 2014.

\bibitem{grother2019face}
Patrick~J Grother, Mei~L Ngan, and Kayee~K Hanaoka.
\newblock Face recognition vendor test part 3: Demographic effects.
\newblock 2019.

\bibitem{huang2017arbitrary}
Xun Huang and Serge Belongie.
\newblock Arbitrary style transfer in real-time with adaptive instance
  normalization.
\newblock In {\em Proceedings of the IEEE International Conference on Computer
  Vision}, pages 1501--1510, 2017.

\bibitem{isola2017image}
P. Isola, J. Zhu, T. Zhou, and A. Efros.
\newblock Image-to-image translation with conditional adversarial networks.
\newblock In {\em Proceedings of the IEEE conference on computer vision and
  pattern recognition}, pages 1125--1134, 2017.

\bibitem{karras2019style}
Tero Karras, Samuli Laine, and Timo Aila.
\newblock A style-based generator architecture for generative adversarial
  networks.
\newblock In {\em Proceedings of the IEEE conference on computer vision and
  pattern recognition}, pages 4401--4410, 2019.

\bibitem{karras2018progressive}
T. Karras, A. Timo, L. Samuli, and L. Jaakko.
\newblock Progressive growing of {GAN}s for improved quality, stability, and
  variation.
\newblock In {\em International Conference on Learning Representations}, 2018.

\bibitem{kim2018DeepVideo}
H. Kim, P. Garrido, A. Tewari, W. Xu, J. Thies, N. Nie{\ss}ner, P. P{\'e}rez,
  C. Richardt, M. Zollh{\"o}fer, and C. Theobalt.
\newblock {Deep Video Portraits}.
\newblock {\em ACM Transactions on Graphics 2018 (TOG)}, 2018.

\bibitem{king2009dlib}
D. King.
\newblock Dlib-ml: {A} machine learning toolkit.
\newblock {\em Journal of Machine Learning Research}, 10(Jul):1755--1758, 2009.

\bibitem{kingma2014adam}
Diederik~P Kingma and Jimmy Ba.
\newblock Adam: A method for stochastic optimization.
\newblock {\em arXiv preprint arXiv:1412.6980}, 2014.

\bibitem{kortylewski2018empirically}
Adam Kortylewski, Bernhard Egger, Andreas Schneider, Thomas Gerig, Andreas
  Morel-Forster, and Thomas Vetter.
\newblock Empirically analyzing the effect of dataset biases on deep face
  recognition systems.
\newblock In {\em Proceedings of the IEEE Conference on Computer Vision and
  Pattern Recognition Workshops}, pages 2093--2102, 2018.

\bibitem{kortylewski2019analyzing}
A. Kortylewski, B. Egger, A. Schneider, T. Gerig, A. Morel-Forster, and T.
  Vetter.
\newblock Analyzing and reducing the damage of dataset bias to face recognition
  with synthetic data.
\newblock In {\em Proceedings of the IEEE Conference on Computer Vision and
  Pattern Recognition Workshops}, pages 2261--2268, 2019.

\bibitem{flame}
T. Li, T Bolkart, M. Black, H. Li, and J. Romero.
\newblock Learning a model of facial shape and expression from {4D} scans.
\newblock {\em ACM Transactions on Graphics}, 36(6):194:1--194:17, Nov. 2017.
\newblock Two first authors contributed equally.

\bibitem{mokhayeri2020cross}
F. Mokhayeri, K. Kamali, and E. Granger.
\newblock Cross-domain face synthesis using a controllable gan.
\newblock In {\em The IEEE Winter Conference on Applications of Computer
  Vision}, pages 252--260, 2020.

\bibitem{nagpal2019deep}
Shruti Nagpal, Maneet Singh, Richa Singh, and Mayank Vatsa.
\newblock Deep learning for face recognition: Pride or prejudiced?
\newblock {\em arXiv preprint arXiv:1904.01219}, 2019.

\bibitem{nguyen2019hologan}
Thu Nguyen-Phuoc, Chuan Li, Lucas Theis, Christian Richardt, and Yong-Liang
  Yang.
\newblock Hologan: Unsupervised learning of 3d representations from natural
  images.
\newblock In {\em Proceedings of the IEEE International Conference on Computer
  Vision}, pages 7588--7597, 2019.

\bibitem{pumarola2018ganimation}
A. Pumarola, A. Agudo, A. Martinez, A. Sanfeliu, and F. Moreno-Noguer.
\newblock Ganimation: Anatomically-aware facial animation from a single image.
\newblock In {\em Proceedings of the European Conference on Computer Vision
  (ECCV)}, pages 818--833, 2018.

\bibitem{coma}
A. Ranjan, T. Bolkart, S. Sanyal, and M. Black.
\newblock Generating 3d faces using convolutional mesh autoencoders.
\newblock In {\em Proceedings of the European Conference on Computer Vision
  (ECCV)}, pages 704--720, 2018.

\bibitem{Robinson_2020_CVPR_Workshops}
J. Robinson, G. Livitz, Y. Henon, C. Qin, Y. Fu, and Samson Timoner.
\newblock Face recognition: Too bias, or not too bias?
\newblock In {\em Proceedings of the IEEE/CVF Conference on Computer Vision and
  Pattern Recognition (CVPR) Workshops}, June 2020.

\bibitem{schroff2015facenet}
Florian Schroff, Dmitry Kalenichenko, and James Philbin.
\newblock Facenet: A unified embedding for face recognition and clustering.
\newblock In {\em Proceedings of the IEEE conference on computer vision and
  pattern recognition}, pages 815--823, 2015.

\bibitem{simonyan2014very}
Karen Simonyan and Andrew Zisserman.
\newblock Very deep convolutional networks for large-scale image recognition.
\newblock {\em arXiv preprint arXiv:1409.1556}, 2014.

\bibitem{szegedy2016inception}
Christian Szegedy, Sergey Ioffe, Vincent Vanhoucke, and Alex Alemi.
\newblock Inception-v4, inception-resnet and the impact of residual connections
  on learning.
\newblock {\em arXiv preprint arXiv:1602.07261}, 2016.

\bibitem{ulyanov2016instance}
Dmitry Ulyanov, Andrea Vedaldi, and Victor Lempitsky.
\newblock Instance normalization: The missing ingredient for fast stylization.
\newblock {\em arXiv preprint arXiv:1607.08022}, 2016.

\bibitem{viazovetskyi2020stylegan2}
Yuri Viazovetskyi, Vladimir Ivashkin, and Evgeny Kashin.
\newblock Stylegan2 distillation for feed-forward image manipulation.
\newblock {\em arXiv preprint arXiv:2003.03581}, 2020.

\bibitem{wang2018video}
T. Wang, M. Liu, J. Zhu, G. Liu, A. Tao, J. Kautz, and B. Catanzaro.
\newblock Video-to-video synthesis.
\newblock In {\em Advances in Neural Information Processing Systems}, pages
  1144--1156, 2018.

\bibitem{wang2018high}
T. Wang, M. Liu, J. Zhu, A. Tao, J. Kautz, and B. Catanzaro.
\newblock High-resolution image synthesis and semantic manipulation with
  conditional gans.
\newblock In {\em Proceedings of the IEEE conference on computer vision and
  pattern recognition}, pages 8798--8807, 2018.

\bibitem{webster2018visual}
B. Webster, S. Kwon, C. Clarizio, S. Anthony, and W. Scheirer.
\newblock Visual psychophysics for making face recognition algorithms more
  explainable.
\newblock In {\em European Conference on Computer Vision}, volume~15, pages
  252--270, 2018.

\bibitem{wiles2018x2face}
O. Wiles, A Sophia~K., and A. Zisserman.
\newblock X2face: A network for controlling face generation using images,
  audio, and pose codes.
\newblock In {\em Proceedings of the European Conference on Computer Vision
  (ECCV)}, pages 670--686, 2018.

\bibitem{zakharov2019few}
E. Zakharov, A. Shysheya, E. Burkov, and V. Lempitsky.
\newblock Few-shot adversarial learning of realistic neural talking head
  models.
\newblock {\em arXiv preprint arXiv:1905.08233}, 2019.

\bibitem{zhenglin:2019:face}
G. Zhenglin, Chen ., and T. Sergey.
\newblock 3d guided fine-grained face manipulation.
\newblock {\em CoRR}, abs/1902.08900, March 2019.

\end{thebibliography}
}

\end{document}